\documentclass[letterpaper, 10 pt, journal, twoside]{IEEEtran}

\usepackage{amsmath,amsfonts}
\usepackage{array}
\usepackage[caption=false,font=footnotesize,labelfont=rm,textfont=rm]{subfig}
\usepackage{textcomp}
\usepackage{stfloats}
\usepackage{url}
\usepackage{verbatim}
\usepackage{graphicx}
\usepackage{cite}
\usepackage{algorithm}
\usepackage{algorithmicx}
\usepackage{algpseudocode}
\usepackage{booktabs}
\usepackage{xcolor}
\usepackage{multirow}
\usepackage{threeparttable}

\usepackage{soul, color, xcolor}
\soulregister{\cite}7 
\soulregister{\citep}7 
\soulregister{\citet}7 
\soulregister{\ref}7 
\soulregister{\pageref}7 

\begin{document}

\title{Multi-Timescale Hierarchical Reinforcement Learning for Unified Behavior and Control of Autonomous Driving}

\author{Guizhe Jin, Zhuoren Li, Bo Leng, Ran Yu, Lu Xiong and Chen Sun

\thanks{Manuscript received: 26 June 2025; Revised 10 September 2025; Accepted 9 October 2025. This paper was recommended for publication by Editor Aniket Bera upon evaluation of the Associate Editor and Reviewers' comments.
This work was supported in part by the National Science Fund for Distinguished Young Scholars of China under Grant No. 52325212, in part by the National Natural Science Foundation of China under Grant No. 52372394, and in part by the Fundamental Research Funds for the Central Universities (22120230311).}

\thanks{Guizhe Jin, Zhuoren Li, Bo Leng, Ran Yu and Lu Xiong are with the School of Automotive Studies, Tongji University, Shanghai 201804, China. (Email: jgz13573016892@163.com, 1911055@tongji.edu.cn, lengbo@tongji.edu.cn, 2433113@tongji.edu.cn, xiong\_lu@tongji.edu.cn).

Chen Sun is with the Department of Data and Systems Engineering, University of Hong Kong, Hong Kong. (Email: c87sun@hku.hk).


This work has been submitted to the IEEE for possible publication. Copyright may be transferred without notice, after which this version may no longer be accessible. 

Digital Object Identifier (DOI): 10.1109/LRA.2025.3623016

}

}

\markboth{IEEE Robotics and Automation Letters. Preprint Version. Accepted October, 2025}
{Jin \MakeLowercase{\textit{et al.}}: Multi-Timescale Hierarchical RL for Autonomous Driving} 


\maketitle

\begin{abstract}
Reinforcement Learning (RL) is increasingly used in autonomous driving (AD) and shows clear advantages. However, most RL-based AD methods overlook policy structure design. An RL policy that only outputs short-timescale vehicle control commands results in fluctuating driving behavior due to fluctuations in network outputs, while one that only outputs long-timescale driving goals cannot achieve unified optimality of driving behavior and control. Therefore, we propose a multi-timescale hierarchical reinforcement learning approach. Our approach adopts a hierarchical policy structure, where high- and low-level RL policies are unified-trained to produce long-timescale motion guidance and short-timescale control commands, respectively. Therein, motion guidance is explicitly represented by hybrid actions to capture multimodal driving behaviors on structured road and support incremental low-level extend-state updates. Additionally, a hierarchical safety mechanism is designed to ensure multi-timescale safety. Evaluation in simulator-based and HighD dataset-based highway multi-lane scenarios demonstrates that our approach significantly improves AD performance, effectively increasing driving efficiency, action consistency and safety.
\end{abstract}


\section{Introduction} \label{sec:Introduction}

\IEEEPARstart{R}{einforcement} learning (RL) has demonstrated strong capabilities in solving sequential decision-making problems, making it a promising paradigm for autonomous driving (AD) applications~\cite{2025RAL_Hu,AI2023_Deng}. However, current RL-based AD approaches often suffer from inappropriate policy output structures, resulting in weak correlations between agent outputs and actual driving behavior. Typically, the RL agent directly outputs vehicle control commands, such as steering angle and acceleration \cite{2022TVT_tang, 2021ITSC_Agarwal}. Fluctuations in policy network’s outputs that arise from its sensitivity to small state variations can cause inconsistent control sequences \cite{2022TVT_Chen}. This makes it difficult to achieve stable and coherent driving, especially in lane-structured scenarios, thereby increasing risks~\cite{2023arXiv_wang, 2024ICDE_xia}.

Hierarchical policy output structures are better suited to AD tasks than directly outputting control commands, as they more closely resemble human driving \cite{2023PartC_Lu}. Behavioral science indicates that human driving behavior is inherently hierarchical in nature, involving both conscious trajectory planning and subconscious action control \cite{2024CSL_Gurses}. Based on this, a common RL approach is to use a hierarchical structure where the high-level policy outputs discrete semantic decisions or trajectory goals, while the low-level rule-based policy generates control commands \cite{2023TVT_Liao}. However, this design limits the flexibility of RL in vehicle control and makes it difficult to produce optimal control commands that adapt to high-level outputs \cite{2022ITSC_Zhao}.

In contrast, a hierarchical structure in which both high- and low-level actions are generated by RL policies better leverages RL’s flexibility, enabling unified learning of driving behaviors and control commands for complex tasks \cite{2025arXiv_zhang}.  Some studies implement this approach using either a single RL agent with a parameterized actor-critic architecture or two independently trained agents \cite{OurITSC, 2022TITS_Peng}. However, these designs often impose timescale consistency constraints on both levels, leading to either fluctuating high-level behaviors or slow low-level control responses \cite{2025arXiv_zhang}. In practice, high-level policy require long-timescale behavioral goals, while low-level policy need short-timescale immediate control \cite{2023Network_chen}. Moreover, given the safety-critical nature of driving, capturing the safer hierarchical policy outputs is also essential.

Therefore, this paper proposes a Multi-Timescale Hierarchical RL approach for autonomous driving. Specifically, we design a hierarchical RL policy structure with high- and low-level components operating at two different timescales: the high-level policy generates long-timescale guidance-actions (i.e., motion guidance), while the low-level policy produces short-timescale execution-actions (i.e., control commands). Both policies are jointly trained to achieve unified optimal performance. Furthermore, we construct a hybrid continuous-discrete action-based motion guidance, enabling multimodal driving behaviors consistent with structured road constraints—discrete laterally and continuous longitudinally. To further enhance safety, we develop a hierarchical safety mechanism that operates in parallel with the policy structure. The main contributions are as follows:
\begin{enumerate}
\item{A multi-timescale hierarchical RL framework is proposed, in which unified-trained policies generate high-level motion guidance and low-level control commands at different timescales, improving driving efficiency while reducing behavior fluctuations.}
\item{A novel motion guidance, explicitly represented by hybrid action, is designed to capture multimodal driving behaviors on structured road. It provides more comprehensive guidance into low-level extend-state through incremental update, enhancing driving performance.}
\item{A supporting hierarchical safety mechanism is proposed, comprising safety evaluation and correction modules, along with a safety-aware termination function. It evaluates motion guidance risks and generates safer high- and low-level actions, enhancing driving safety across different timescales.}
\end{enumerate}


\section{Related Works}\label{sec:Related Works}


A general RL-based AD approach with a hierarchical architecture combines RL with a rule-based method for vehicle control. Specifically, RL is used for the high-level policy, producing outputs that can either be semantic decisions (e.g., lane changes) \cite{2024TITS_Lee, 2023TVT_wang, 2024ICDE_xia, 2023TVT_Liao}, motion primitives from a discrete space \cite{2023PartC_Lu, 2024AIAS_jin, 2022AI_Lu}, or target points in a continuous space \cite{2023IROS_zhou, 2023arXiv_wang, 2021NeurIPS_Dalal, 2025RAL_Chen}. Based on these behavior goals, the low-level rule-based controller generates actual vehicle control commands (e.g., steering angle, acceleration). However, this structure limits the flexibility of the RL policy due to indirect vehicle control. Additionally, the low-level controller may fail to respond effectively to dynamic environmental changes, or its response may deviate from the intended high-level behavior, preventing unified optimization across both levels \cite{2022ITSC_Zhao}.

In contrast, using RL policies to simultaneously generate both abstract driving behaviors and concrete control commands is more advanced. Some classical studies construct implicit hierarchical policy structures \cite{2022TVT_Chen, OurITSC, 2024Mach_Lin, 2025arXiv_jin, 2024SAE_li, 2024TITS_chen}, or train independent RL agents for high- and low-level policies \cite{2022ITSC_Zhao,2025arXiv_zhang,2022TITS_Peng,2021PartC_Guo}, to enhance unified optimization between the two levels. However, these methods face timescale consistency constraints, making it difficult to set appropriate timescales for both levels. Specifically, a too-short timescale leads to driving behavior fluctuations, while a too-long timescale slows responses to dynamic environment changes.

Further, some studies attempt to use RL with different timescales to construct hierarchical policy structures, commonly adopting a skill-based approach \cite{2024CSL_Gurses, 2024PartC_Mao, 2023ICRA_Lee}. For such an approach, low-level RL agents are pre-trained to output sequences of control commands over short timescales, known as motion skills. A high-level RL agent is then trained to select the optimal motion skill from this skill space. This approach breaks the timescale consistency constraint between different levels, thereby better leveraging the flexibility of RL. However, the skill space is typically fixed, and the high-level RL policy essentially learns to combine these time-extended control commands. This limits the potential of RL to explore optimal actions at each driving step \cite{2024RAL_Hao}.

\begin{figure}[!tb]
  \centering
  \includegraphics[width=0.48\textwidth]{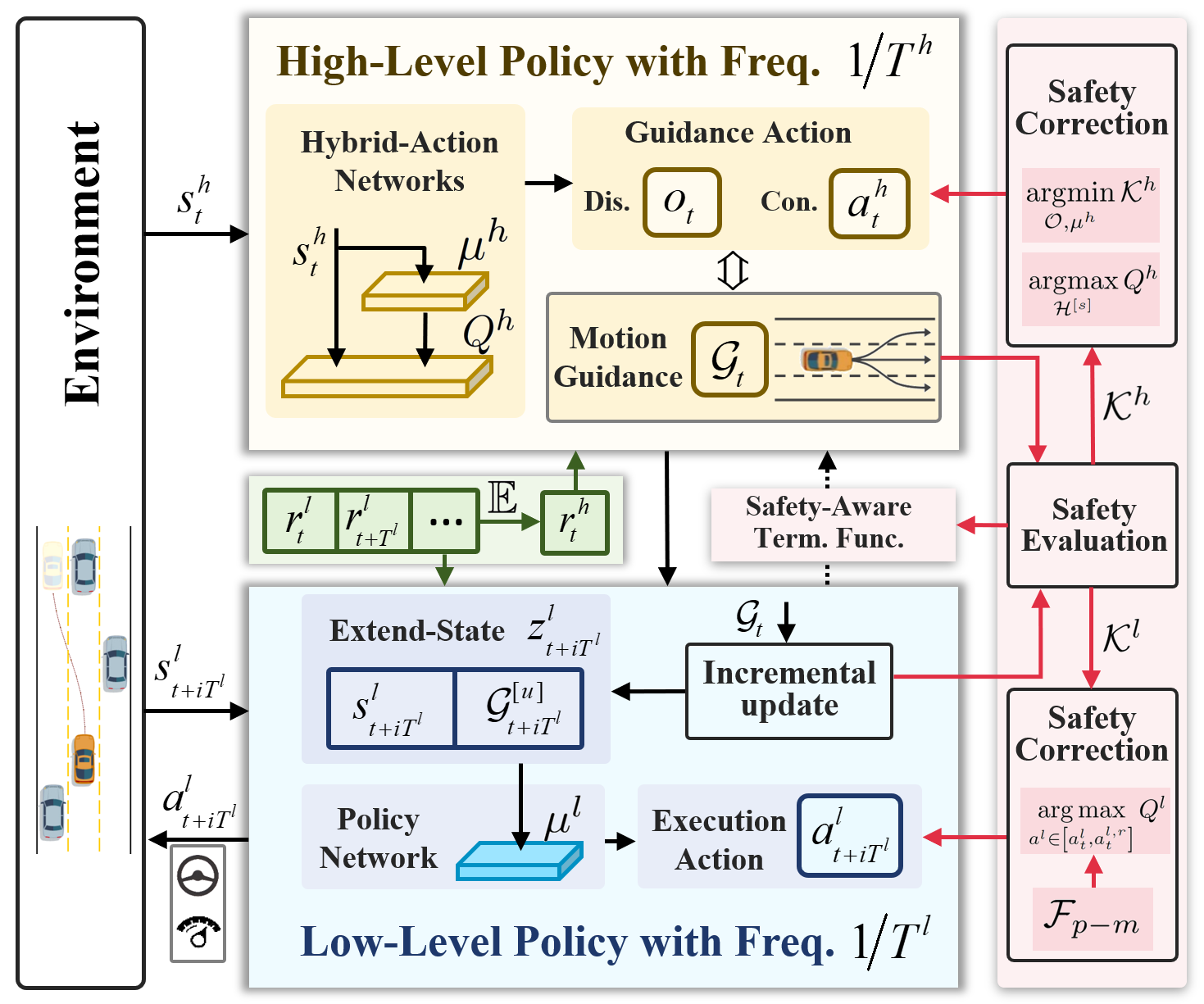}
  \vspace{-2mm}
  \caption{The framework of multi-timescale hierarchical RL approach. It consists of two unified-trained policies at different levels, with supporting safety mechanisms. High‑level motion guidance that, combined with environment features, forms the extended input of the low‑level policy, while low‑level rewards are fed back in expectation to the high‑level for joint optimization.}
  \label{framework}
\vspace{-2mm}
\end{figure}

To address the limitations of previous works, we propose a multi-timescale hierarchical RL approach. Two joint-trained hierarchical RL policies output long-timescale abstract motion guidance and short-timescale concrete control commands, respectively. Few studies have explored similar approaches in AD \cite{2023Network_chen, 2024RAL_Hao}, and those that do have notable shortcomings: (1) high-level outputs are restricted to either purely discrete or continuous action spaces, failing to match structured road constraints; and (2) hierarchical policies lack safety considerations. In contrast, we integrate the parameterized actor-critic (P-AC) technique into the hierarchical structure, explicitly representing high-level motion guidance with discrete-continuous hybrid actions. Additionally, a hierarchical safety mechanism is designed to support the policy structure. The framework of our approach is illustrated in Fig.~\ref{framework}.

\section{Methodology}\label{sec:Methodology}

\subsection{Framework Construction}\label{sec:Methodology:A}

\subsubsection{MDP Re-Formulation}\label{sec:Methodology:A:1}

Inspired by DAC \cite{2019NeurIPS_zhang}, we reformulate the training as two augmented Markov Decision Processes (MDPs): high-MDP ${\cal M}^h$ and low-MDP ${\cal M}^l$. The high-level policy $\pi^h$ and low-level policy $\pi^l$ make decisions within their respective MDPs and are optimized jointly.

The ${\cal M}^h$ can be defined by a tuple $<{{\cal S}^h},{{\cal H}^h},{{\cal R}^h},{{\cal T}^h},\gamma >$, where: 1) ${\cal S}^h$ is the high-level state space, derived from the environment; 2) ${\cal H}^h$ is the hybrid action space, composed of discrete and continuous subspaces, ${\cal O}$ and ${\cal A}^h$; 3) ${\cal R}^h$ is the high-level reward function, determined by low-level rewards and agent violations; and 4) ${\cal T}^h$ and $\gamma$ are the high-level transition function and discount factor.

Similarly, the ${\cal M}^l$ is defined by $<{{\cal Z}^l},{{\cal A}^l},{{\cal R}^l},{{\cal T}^l},\gamma>$, where: 1) ${\cal Z}^l$ is the low-level state space, combining the original low-level state space ${\cal S}^l$ and all motion guidance mapping to ${\cal H}^l$; 2) ${\cal A}^l$ is the low-level continuous action space; 3) ${\cal R}^l$ is the low-level reward function derived from environmental feedback; and 4) ${\cal T}^l$ is the transition function.

At each long timestep ${T^h}$, ${\pi ^h}( o, {a^h} \mid {s^h} )$ outputs guidance-action $(o, {a^h}) \in {{\cal H}^h}$, with ${s^h} \in {{\cal S}^h}$. Each guidance-action explicitly represents a motion guidance ${\cal G}$ via a bi-directional mapping: ${\cal G} \leftrightarrow (o, {a^h})$. Hereafter, ${\cal G}(o, a^h)$ denotes the motion guidance corresponding to $(o, a^h)$. Then, at each short timestep ${T^l}$, ${\pi ^l}( {a^l}\! \mid \!{z^l} )$ receives extend-state ${z^l} = ({s^l}, {\cal G}(o, {a^h}))$ from the environment and the high-level policy $\pi^h$, where ${z^l} \in {{\cal Z}^l}$ and ${s^l} \in {{\cal S}^l}$, to output a execution-action $a^l \in {{\cal A}^l}$. The timesteps are related by ${T^h} = n{T^l}$, where $n$ is determined by the termination function $\beta$, i.e., ${\arg _i}\left[ {\beta ( {z_{t + i{T^l}}^l} ) = 1} \right]$. Here, $z_{t + i{T^l}}^l$ is required for the $i$-th output of ${\pi ^l}$ after receiving $\cal G$. Thus, each high-level motion guidance corresponds to a variable-length sequence of low-level control commands.

\subsubsection{Safety Mechanism Definition}\label{sec:Methodology:A:2}

The safety mechanism for the hierarchical policy includes: 1) safety evaluation module, 2) high- and low-level independent safety correction module, and 3) the safety-aware termination function. The safety evaluation module crosses both levels and generates risk severity of motion guidance ${\cal K}( {{\cal G}( {o,{a^h}} ),s} )$ at different timescales. The safety correction module fuses ${\cal K}$ with action values to produce safer actions $o^{[s]}$, $a^{h,[s]}$, and $a^{l,[s]}$. The safety-aware termination function $\beta$ combines the safety evaluation results from both levels, prioritizing high-level correction to further enhance safety.

\subsection{Multi-Timescale Hierarchical Policy Design}\label{sec:Methodology:B}

The policies $\pi^h$ and $\pi^l$ operate in two parallel augmented MDPs, which is trained simultaneously under same sampling conditions \cite{2019NeurIPS_zhang}. Optimizing $\pi^h$ requires the parameterized actor-critic algorithm \cite{2018arXiv_Xiong}, while any policy optimization algorithm can be applied to $\pi^l$. To achieve joint optimality of $\pi^h$ and $\pi^l$, a strong coupling between the two policies is essential: 1) The guidance-action from $\pi^h$ is incorporated into the extend-state as input to $\pi^l$, and 2) the rewards obtained by $\pi^l$ within the timestep ${T}^l$ are also used to update $\pi^h$.

\subsubsection{High-Level Policy}\label{sec:Methodology:B:1}

In high-MDP, the state-action value function for the optimal high-level policy is defined by the following Bellman optimality equation:
\begin{equation}
\label{Q^h}
\begin{split}
{Q^h}&\left( {s_t^h,{o_t},a_t^h} \right) = \\ &{\mathbb{E}} \left[ {r_{t + {T^h}}^h + \gamma \mathop {\max }\limits_{o \in {\cal O}} \mathop {\sup }\limits_{{a^h} \in {{\cal A}^h}} {Q^h}\left( {s_{t + {T^h}}^h,o,{a^h}} \right)} \right],
\end{split}
\end{equation}
where $r_{t + {T^h}}^h \in {\cal R}^h$ is given by:
\begin{equation}
\label{r^h}
\begin{split}
r_{t + {T^h}}^h = \left( {1 - {f_{v}}\left( {{s^l}} \right)} \right){{\mathbb{E}}_{i = 1 \sim n}}\left[ {r_{t + i{T^l}}^l} \right] + {f_{v}}\left( {{s^l}} \right){{\cal R}_{vio}},
\end{split}
\end{equation}
where $r_{t + i{T^l}}^l \in {\cal R}^l$ is the feedback reward from the environment for the $i$-th action of $\pi^l$. The violation flag function $f_v$ is set to 1 in case of agent violations (e.g., vehicle collisions) and 0 otherwise. Introducing $f_v$ prevents the dilution of violation-related rewards ${\cal R}_{vio}$ by expectation-seeking operations, ensuring that the high-level policy maintains a strong emphasis on violations. The definition of ${\cal R}_{vio}$ is provided in Sec.~\ref{sec:Implementation:A}.

However, finding the optimal $a^h$ in a hybrid action space is challenging. Following the idea of parameterized actor-critic, the high-level policy $\pi^h$ outputs $a^h$ through the cooperation between a deterministic policy network $\mu^h(s^h; \theta^h)$ and a value network $Q^h(s^h, o, a^h; \omega^h)$. Details of this cooperation can be found in our previous work~\cite{OurITSC}. Thus, with ${\pi ^h} = ( { \cdot \left| {{\mu ^h}( { \cdot {\rm{ }};{\theta ^h}} ),{Q^h}( { \cdot {\rm{ }};{\omega ^h}} ),{s^h}} \right.} )$, the optimal state-action value function can be rewritten as:
\begin{equation}
\label{Q^h_2}
\begin{split}
&{Q^h} \left( {s_t^h,{o_t},a_t^h} \right) = \\ & {\mathbb{E}} \left[ {r_{t + {T^h}}^h + \gamma \mathop {\max }\limits_{o \in {\cal O}} {Q^h}\left( {s_{t + {T^h}}^h,o,{\mu ^h}\left( {s_{t + {T^h}}^h;{\theta ^h}} \right);{\omega ^h}} \right)} \right].
\end{split}
\end{equation}
This function’s solution is the optimal guidance-action $(o, a^h)$.

The $(o, a^h)$ explicitly represents the motion guidance ${\cal G}$ through a bi-directional mapping:
\begin{equation}
\label{a_G}
\begin{split}
{\cal G} \leftarrow \Psi \left( {o,{a^h}} \right),{\rm{  }}\left( {o,{a^h}} \right) \leftarrow {\Psi ^{ - 1}}\left( {\cal G} \right).
\end{split}
\end{equation}
where $\Psi$ is an explicit representation function, depending on the practical significance of $(o, a^h)$ and $\cal G$. A generalized example for AD is provided in Sec.~\ref{sec:Implementation:A}.

\subsubsection{Low-Level Policy}\label{sec:Methodology:B:2}

\begin{figure}[!tb]
  \centering
  \includegraphics[width=0.4\textwidth]{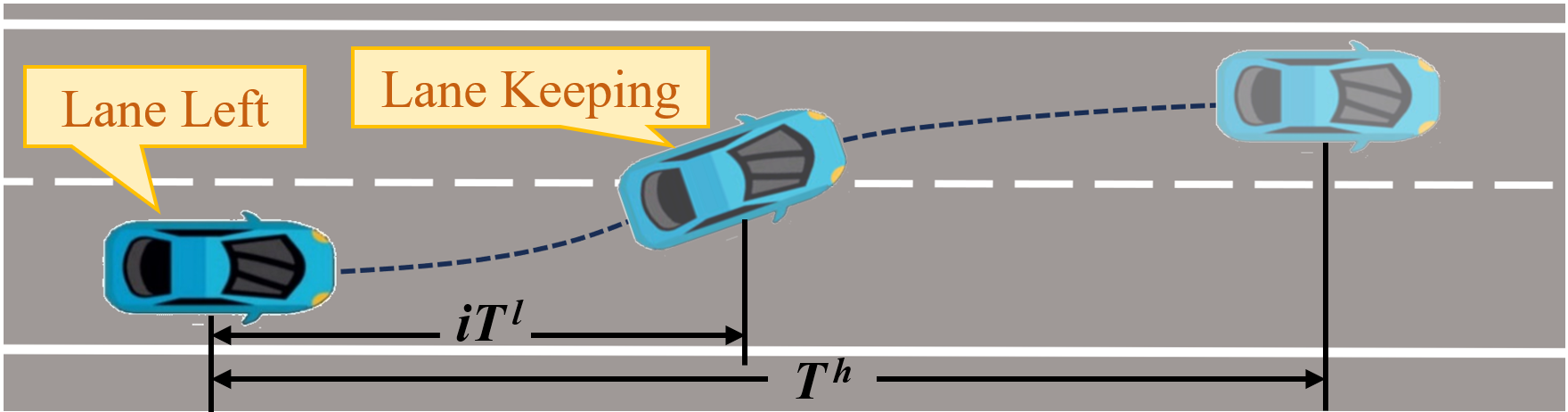}
  \vspace{-2mm}
  \caption{Illustration of the low-level extend-state transition. Assume that $\pi ^ h$ generates a guidance-action of 'Lane Left' for $T^h$. When the vehicle crosses the lane divider at $iT^l$, the guidance-action observed from the agent's viewpoint becomes 'Lane Keeping', even though the $T^h$ has not yet ended.}
  \label{LC_example}
\vspace{-2mm}
\end{figure} 

In low-MDP, ${\cal G}(o, a^h)$ is provided every $T^h$, while the low-level policy $\pi^l$ acquires the observed extend-state $z_{t + i{T^l}}^l$ at much shorter timestep $T^l$. As a result, the high-level output observed from the low-level perspective may change, as illustrated in Figure~\ref{LC_example} with a lane-change scenario. Directly using the fixed ${\cal G}(o, a^h)$ in constructing $z_{t + i{T^l}}^l$ would introduce state inconsistencies, hindering stable training of $\pi^l$. To address this, motion guidance is incrementally updated at each short timestep using physical information, allowing the guidance-action to be naturally updated:
\begin{equation}
\label{real==}
\begin{split}
\left(\! {o_{t + i{T^l}}^{\left[ u \right]},a_{t + i{T^l}}^{h,\left[ u \right]}}\! \right)\! \stackrel{\Phi}{\leftrightarrow }\! {{\cal G}_{t + i{T^l}}^{\left[ u \right]}}\! \leftarrow \!{\cal G}_{t}\!\left( {{o_t},a_t^h} \right)\!, \forall i \!\in \!\left[ {1,\! \cdots \!,n} \right],
\end{split}
\end{equation}
where the superscript $u$ denotes the variable has undergone incremental updating. Accordingly, the actual low-level extend-state becomes $z_{t + i{T^l}}^l = ( {s_{t + i{T^l}}^l,{\cal G}_{t + i{T^l}}^{\left[ u \right]}} )$. Since motion guidance is environment-specific, incremental updating using physical information is straightforward to implement. An example is provided in Sec.~\ref{sec:Implementation:A}.

Therefore, in low-MDP, the state-action value function for the optimal low-level policy is given by the following Bellman optimality equation:
\begin{equation}
\label{low-MDP-1}
\begin{split}
{Q^l} \left( {z_t^l,a_t^l} \right) = \mathbb{E} \left[ {r_{t + {T^l}}^l + \gamma U\left( {z_{t + {T^l}}^l} \right)} \right],
\end{split}
\end{equation}
\begin{equation}
\label{low-MDP-2}
\begin{split}
U \left( {z_{t + {T^l}}^l} \right) =& \left( {1 - \beta \left( {s_{t + {T^l}}^l} \right)} \right)\mathop {\sup }\limits_{{a^l} \in {{\cal A}^l}} {Q^l}\left( {z_{t + {T^l}}^l,{a^l}} \right) + \\
&\beta \left( {s_{t + {T^l}}^l} \right)\mathop {\sup }\limits_{{a^l} \in {{\cal A}^l}} {Q^l}\left( {{\pi ^h}\left( {s^h} \right),s_{t + {T^l}}^l,{a^l}} \right),
\end{split}
\end{equation}
where $U( {z_{t + {T^l}}^l} )$ is the optimal state value function. Since $\pi^l$ outputs continuous actions and is compatible with any optimization algorithm, it can be directly approximated using a policy network ${\mu ^l}( {{z^l}{\rm{ }};{\theta ^l}} )$. Meanwhile, a value network ${Q^l}( {z_t^l,a_t^l;{\omega ^l}} )$ is introduced to estimate the state-action value. 

\subsection{Hierarchical Safety Mechanism}\label{sec:Methodology:C}


The original motion guidance from $\pi^h$ may pose safety risks. To quantify these risks, a safety evaluation module is used to assess risk severity: ${\cal K}_t^h( {{\cal G}_t( {{o_t},a_t^h} ),s_t^h} )$. Since the parameterized actor-critic provides values for all alternative actions, when risk exceeds a threshold, i.e., $\eta {\cal K}_t^h \ge {{\cal K}_{th}}$, the safety correction module generates safer guidance-actions guided by ${{\cal K}^h}$ and $Q^h$:
\begin{equation}
\label{high_s_action}
\begin{split}
\begin{array}{l}
\left(\! {o_t^{\left[ s \right]},\!a_t^{h,\left[ s \right]}}\! \right)\! =\! \left\{{\begin{array}{*{2}{l}}\!
{\mathop {\arg\! \max }\limits_{{{\cal H}^{[s]}}} {Q^h}\!\left(\! {s_t^h,\!o,\!{a^h};\!{\omega ^h}\!} \right)}&{{{\cal H}^{\left[ s \right]}} \!\ne\! \emptyset }\\
\!{\mathop {\arg\! \min }\limits_{{\cal O}, {\mu ^h}} {{\cal K}^h}\!\left( {{\cal G}\!\left( \!{o,\!{a^h}}\! \right)\!,\!s_t^h} \right)}&{{{\cal H}^{\left[ s \right]}}\! =\! \emptyset }
\end{array}} \right.\\
\end{array}
\end{split}
\end{equation}
\begin{equation}
\label{H^h}
\begin{split}
\begin{array}{l}
{{\cal H}^{\left[ s \right]}}\! =\! \mathop {\arg }\limits_ {o,{a^h}}\! \left[ {{\eta {\cal K}^h}\!\left( {{\cal G}\left( {o,{a^h}} \right),s_t^h} \right)\! <\! {{\cal K}_{th}}} \right]\!\left| {_{\forall o \in {\cal O},{a^h} = {\mu ^h}}} \right.
\end{array}
\end{split}
\end{equation}
where the superscript $[s]$ indicates processing by the safety mechanism. The variable $\eta$ is an attention weight for ${\cal K}_t^h$, gradually increased during training to avoid early convergence to a conservative policy. The ${{\cal H}^{\left[ s \right]}}$ represents a safe guidance-action space comprising alternative actions with risk severity below ${{\cal K}_{th}}$. Then, safer motion guidance ${{\cal G}_t^{\left[ s \right]}}$ is reconstructed based on $( {o_t^{\left[ s \right]},a_t^{h,\left[ s \right]}} )$, and they are updated to ${{\cal G}_t^{\left[ {s,u} \right]}}$ and $( {o_t^{\left[ {s,u} \right]},a_t^{h,\left[ {s,u} \right]}} )$ according to Eq.~\ref{real==}. Additionally, $( {o_t^{\left[ s \right]},a_t^{h,\left[ s \right]}} )$ forms a tuple with $s_t^h$, $r_t^h$, and $s_{t+T^h}^h$, which is stored in the high-level replay buffer ${\cal D}^h$ for updating $\pi^h$.


The safety mechanism for $\pi^h$ cannot always ensure low risk severity, as execution-actions from $\pi^l$ or dynamic environmental changes may lead to sudden risk increases. To address this, the safety evaluation module operating on a shorter timescale is introduced to assess risk severity: ${\cal K}_t^l( {\cal G}_t^{[s,u]}( o_t^{[s,u]}, a_t^{h,[s,u]} ), s_t^l )$. Since $\pi^l$ outputs a deterministic action, an alternative action $a_t^{l,r}$ is obtained from a priori conservative control model ${{\cal F}_{p-m}}$, to generate a safer action:
\begin{equation}
\label{low_s_a}
\begin{split}
\begin{array}{l}
a_t^{l,\left[ s \right]} = \left\{ {\begin{array}{*{20}{l}}
{a_t^l}&{\eta {\cal K}_t^l < {{\cal K}_{th}}}\\
{\mathop {\arg \max }\limits_{{a^l} \in \left[ {a_t^l,a_t^{l,r}} \right]} {Q^l}\left( {z_t^l,{a^l};{\omega ^l}} \right)}&{\eta {\cal K}_t^l \ge {{\cal K}_{th}}}
\end{array}} \right.
\end{array}
\end{split}
\end{equation}
where ${{\cal F}_{p - m}}$ may vary depending on the specific implementation, with details and an example provided in Sec.~\ref{sec:Implementation:B}. The $a_t^{l,\left[ s \right]}$ is used to control the agent’s interaction with the environment and, together with $z_t^l$, $r_t^l$, and $z_{t + {T^l}}^l$, is stored in the low-level replay buffer ${\cal D}^l$ for updating $\pi^l$.


In fact, the high- and low-level safety mechanisms are not independent; they cooperate within a unified framework to safeguard both long‑term behaviour and short‑term control. When the low‑level mechanism detects imminent risk of motion guidance, the high-level mechanism should be prioritized over a longer timescale to ensure longer-term safety. Accordingly, the safety evaluations results of $\pi^h$ and $\pi^l$ are integrated, resulting in a safety-aware termination function $\beta$:
\begin{equation}
\label{beta}
\begin{split}
\begin{array}{l}
\beta\! \left(\! {z_{t + i{T^l}}^l}\! \right)\! =\! {f_{v}}\!\left(\! {s_{t + i{T^l}}^l} \!\right)\! \vee\! \left[ {i\! =\! {n_{\max }}} \right]\! \vee\! {\cal C}\left(\! {z_{t + i{T^l}}^l}\! \right)
\end{array}
\end{split}
\end{equation}
\begin{equation}
\label{C}
\begin{split}
\begin{array}{l}
{\cal C}\left( {z_{t + i{T^l}}^l} \right) = \left[ {\left( {\eta {\cal K}_{t + i{T^l}}^l \ge {{\cal K}_{th}}} \right) \wedge \left( \eta {\cal K}_t^h < {{\cal K}_{th}} \right)} \right]
\end{array}
\end{split}
\end{equation}
Three conditions trigger $\beta ( {z_{t + i{T^l}}^l} ) = 1$: 1) the agent is in violation, i.e., ${f_v} = 1$; 2) the cumulative number of low-level decisions reaches the limit ${n_{\max }}$, which can be fixed or task-dependent; and 3) ${\cal C}( z_{t + i{T^l}}^l ) = 1$, indicating that, during the operation of $\pi^l$, the risk severity exceeds the threshold while the safety correction of $\pi^h$ is inactive.

\subsection{Policy Optimization Training}\label{sec:Methodology:D}

The update of $\pi^h$ involves two networks: ${Q^h}(\cdot\,;{\omega^h})$ and ${\mu^h}(\cdot\,;{\theta^h})$. Corresponding target networks, ${Q^h_*}(\cdot\,;{\omega^h})$ and ${\mu^h_*}(\cdot\,;{\theta^h})$, are introduced, being updated using a soft update parameter $\tau$. The gradient for updating ${Q^h}(\cdot\,;{\omega^h})$ is computed from randomly sampled transitions $<\! s_t^h, o_t^{[s]}, a_t^{h,[s]}, r_{t+T^h}^h, s_{t+T^h}^h \! >$ from ${\cal D}^h$, and is given by:
\begin{equation}
\label{L_Q_h}
\begin{split}
\begin{array}{l}
{\nabla}\!{{\cal L}_t}\!\left( \!{{\omega ^h}} \!\right)\! =\!  - \!\left[ {y\!\left(\! {{Q^h}}\! \right)\! -\! {Q^h}\!\left(\! {s_t^h,\!o_t^{\left[ s \right]}\!,\!a_t^{h,\left[ s \right]};\!{\omega ^h}}\! \right)}\! \right]\!{\nabla \!_{{\omega ^h}}}{Q^h}\\
y\!\left(\! {{Q^h}}\! \right)\! =\! r\!_{t \!+ \!{T^h}}^h\! +\! \gamma\! \mathop {\max }\limits_{o \in {\cal O}}\! Q\!_*^h\!\left(\! {s\!_{t\! +\! {T^h}}^h,\!o,\!\mu _*^h\!\left(\! {s\!_{t\! + \!{T^h}}^h\!;\!\theta _*^h}\! \right)\!;\!\omega _*^h}\! \right)\!.
\end{array}
\end{split}
\end{equation}
For ${\mu ^h}( { \cdot {\rm{ }};{\theta ^h}} )$, the update objective is to maximize the value function over all discrete actions, so its policy gradient is:
\begin{equation}
\label{J_u_h}
\begin{split}
{\nabla}\!{{\cal J}_t}\!\left(\! {{\theta ^h}}\! \right)\! =\! \sum\limits_{o \in {\cal O}}\! {{\nabla\! _{{\theta ^h}}}{\mu ^h}\!\left(\! {s_t^h;\!{\theta ^h}}\! \right)\!{\nabla\! _{{a^h}}}{Q\!^h}\!\left(\! {s_t^h,\!o,\!{a\!^h};\!{\omega ^h}}\! \right)} \!\left|\! {_{{a^h}\! =\! {\mu ^h}}} \right.
\end{split}
\end{equation}

Similarly, updating the low-level ${Q^l}( { \cdot {\rm{ }};{\omega ^l}} )$ and ${\mu ^l}( { \cdot {\rm{ }};{\theta ^l}} )$ relies on target networks $Q_*^l( { \cdot {\rm{ }};\omega _*^l} )$ and $\mu _*^l( { \cdot {\rm{ }};\theta _*^l} )$. The gradient of ${Q^l}( { \cdot {\rm{ }};{\omega ^l}} )$ is computed based on $ < z_t^l,a_t^{l,\left[ s \right]},r_{t + {T^l}}^l,z_{t + {T^l}}^l > $, which is randomly sampled from ${\cal D}^l$:
\begin{equation}
\label{L_Q_l}
\begin{split}
\begin{array}{l}
\nabla {{\cal L}_t}\left( {{\omega ^l}} \right) =  - \left[ {y\left( {{Q^l}} \right) - {Q^l}\left( {z_t^l,a_t^{l,\left[ s \right]};{\omega ^l}} \right)} \right]{\nabla _{{\omega ^l}}}{Q^l},\\
y\left( {{Q^l}} \right) = r_{t + {T^l}}^l + \gamma U\left( {z_{t + {T^l}}^l} \right)\left| {_{\pi _*^h,{s^h} = s_{t + {T^l}}^l}} \right.,
\end{array}
\end{split}
\end{equation}
where $\pi _*^h$ denotes the high-level target policy. The gradient for updating ${\mu ^l}( { \cdot {\rm{ }};{\theta ^l}} )$ is:
\begin{equation}
\label{J_u_l}
\begin{split}
\begin{array}{l}
\nabla {{\cal J}_t}\left( {{\theta ^l}} \right) = {\nabla _{{\theta ^{_l}}}}{\mu ^l}\left( {z_t^l;{\theta ^l}} \right){\nabla _{{a^l}}}{Q^l}\left( {z_t^l,{a^l};{\omega ^l}} \right)\left| {_{{a^l} = {\mu ^l}}} \right..
\end{array}
\end{split}
\end{equation}

The training procedure of our multi-timescale hierarchical RL, with safety mechanisms, is presented in Algorithm~\ref{alg1}.

\begin{algorithm}[!tb]
    \caption{Training process of our method} \label{alg1}
    \begin{algorithmic}[1] 
    \Require total training steps $T$, $T^h$, $T^l$, $\{\tau\}$, ${\cal K}_{th}$, $n_{\max }$.
        \State Initialize: replay buffer $\{{\cal D}^h, {\cal D}^l\}$, high- and low-level networks $\{Q^h, \mu^h, Q_{*}^h, \mu _{*}^h, Q^l, \mu^l, Q_{*}^l, \mu _{*}^l\}$.
        \For{$t = 0$ to $T$} 
            \State Get ${s_t^h}$ from environment.
            \State Select $a_t^h \sim {\mu ^h}\left( {{s^h_t};{\theta ^h}} \right)$ for ${\forall o \in {\cal O}}$.
            \State Select $\left( {{o_t},a_t^h} \right) \sim { {Q^h}\left( {s_{t}^h,o, a_t^h;{\omega ^h}} \right)} \left| {_{\forall o \in {\cal O}}} \right.$. 
            \State Construct motion guidance ${\cal G}_t$ according to $\Psi \left(  {{o_t},a_t^h} \right)$.
            \State Get safer $\left( {o_t^{\left[ s \right]},a_t^{h,\left[ s \right]}} \right)$ and ${{\cal G}_t^{\left[ s \right]}}$ according to Eq.~\ref{high_s_action}.
            \While{ not $\beta$} 
                \State Select $a_t^l \sim {\mu ^l}\left( {{z^l_t}{\rm{ }};{\theta ^l}} \right)$.
                \State Get safer $a_t^{l,\left[ s \right]}$ according to Eq.~\ref{low_s_a}.
                \State Get $s^l_{t + T^l}$ and $r^l_{t + T^l}$ from environment.
                \State Incremental update for ${\cal G}_{t+T^l}^{\left[ {s,u} \right]}\left( {o_{t+T^l}^{\left[ {s,u} \right]},a_{t+T^l}^{h,\left[ {s,u} \right]}} \right)$.
                \State Get $\beta \left( {z_{t+T^l}^l} \right)$ according to Eq.~\ref{beta}.
                \State Update $\omega ^l_{t + T^l}$, $\theta ^l_{t + T^l}$, ${\omega ^{l}_{*}}$, ${\theta ^{l}_{*}}$.
                \State $z _{t}^l \leftarrow z _{t + T^l}^l$, $i \leftarrow i + 1$, $t \leftarrow t + T^l$.
            \EndWhile
            \State $n \leftarrow i$, $T^h = nT^l$, $s _{t + T^h}^h \leftarrow s _{t + T^l}^l$.
            \State Get ${r_{t + {T^h}}^h}$ according to Eq.~\ref{r^h}.
            \State Update $\omega ^h_{t + T^h}$, $\theta ^h_{t + T^h}$, ${\omega ^{h}_{*}}$, ${\theta ^{h}_{*}}$.
            \State $s_t^h \leftarrow s_{t + T^h}^h$, $i \leftarrow 0$, $\beta \leftarrow 0$.
        \EndFor
    \end{algorithmic}
\end{algorithm}

\section{Implementation}\label{sec:Implementation}

The highway multi-lane scenario is a common yet challenging environment, requiring the ego vehicle (EV) to dynamically adjust its position and speed within defined lanes to ensure efficiency, action consistency, and safety. Therefore, we implement our approach in this setting.

\subsection{Two Augmented MDPs Formulation}\label{sec:Implementation:A}

Both high-MDP and low-MDP involve three key elements: 1) Action space: The high-MDP adopts a hybrid action space for motion guidance under road constraints, focusing on long-term target position planning. In contrast, low-MDP uses a continuous action space to generate control commands, focusing on short-term speed adjustment. 2) State space: The two MDPs share a same original state space, i.e., ${\cal S}^h = {\cal S}^l$. 3) Reward function: The high-MDP’s reward is implicitly defined by that of the low-MDP and requires no separate design.

\subsubsection{High-MDP Action Space}

In lanes that are discrete laterally but continuous longitudinally, the hybrid action space ${\cal H}^h$ of the high-level policy is defined as follows:
\begin{equation}
\label{O_and_A^h}
\begin{split}
\begin{array}{l}
\left\{ \begin{array}{l}
{\cal O}: \left\{ { - {w_r},0,{w_r}} \right\}\\
{{\cal A}^h}: \left[ {\min \left( {\sqrt {4{R_0}{w_r} - w_r^2} ,\frac{{v_e^2}}{{2a_{\max }^ - }}} \right),{e^{\left| {{v_e}} \right| + {w_r}}}} \right]
\end{array} \right.
\end{array} 
\end{split}
\end{equation}
where $w_r$ is the lane width, ${R_0}$ is the minimum turning radius, ${2a_{\max }^-}$ is the maximum braking acceleration, and $v_e$ is the EV's speed. The ${\cal O}$ allows $o$ to represent lane selection, restricting the target to the current or adjacent lane. Given the EV's state and kinematics, the ${\cal A}^h$ allows $a^h$ to represent the selection of a feasible target location within the chosen lane. Together, $(o, a^h)$ specifies a target point without further processing, which can be considered as a coarse-grained motion guidance.

To provide finer-grained guidance for the low-level policy, the motion guidance is represented as a series of path points, which are generated from a polynomial curve: ${\cal G} = {\arg _{(x_j, y_j)}}[ y_j = \sum_{m=0}^5 c_m x_j^m ]$, where $j \in [1, \cdots, g]$. Here, $\cal G$ is a set of $g$ target points lying on a fifth-degree polynomial parameterized by coefficients $c_m$. In the Frenet frame, with the EV at the origin, $x_j$ and $y_j$ are the longitudinal and lateral coordinates of the $j$-th target point. The start point is set by the EV’s current state: $(x_1, y_1) = (x_e, y_e)$, with heading angle $\varphi_1 = \varphi_e$. The end point is given by $\pi^h$: $(x_g, y_g) = (o, a^h)$, with $\varphi_g$ from lane information. Solving the resulting linear system yields the coefficients $c_m$ \cite{OurITSC}. The mapping $(o, a^h) \equiv (x_g, y_g) \in {\cal G}$ defines an explicit representation function $\Psi$, which satisfies Eq.~\ref{a_G} given the current EV and road states.

In timestep $T^h$, the frenet frame moves with EV. Since the points in $\cal G$ are defined relative to previous frenet frame, their coordinates must be updated. A simple coordinate transformation yields the updated set ${{\cal G}^{[u]}}$, corresponding to Eq.~\ref{real==}.

\subsubsection{Low-MDP Action Space}

The control commands required for EV driving are acceleration and steering angle. Thus, the execution action output by $\pi^l$ consists of two items: $a^l = (\delta_e, a_e)$. According to general vehicle kinematics, the low-level action space is defined as: 
\begin{equation}
\label{A^l}
\begin{split}
{\cal A}^l = \left\{\left[-\pi/6 \text{ rad}, \pi/6 \text{ rad}\right], \left[-3\text{ m/s}^2,3\text{ m/s}^2\right]\right\}.
\end{split}
\end{equation}

\subsubsection{State Space}

Both $\pi^h$ and $\pi^l$ should account for the states of the EV and surrounding vehicles (SVs) in adjacent lanes. Thus, the state space is defined as:
\begin{equation}
\label{state_space}
\begin{split}
{{\cal S}^h}\! \equiv\! {{\cal S}^l}\! =\! \left\{\! {\begin{array}{*{20}{c}}
{{{\left[ {I{D_{lane}},{x_e},{y_e},{\varphi _e},v_e^x,v_e^y} \right]}^{{\rm{EV}}}},}\\
{\left[ {{p_k},\Delta {x_k},\Delta {y_k},{\varphi _k},\Delta v_k^x,\Delta v_k^y} \right]_{k \in \left[ {1 \cdots 6} \right]}^{{\rm{SVs}}}}
\end{array}}\! \right\}\!,
\end{split}
\end{equation}
The EV state includes lane ID, longitudinal and lateral positions, heading angle, and longitudinal and lateral speeds. SVs ahead of and behind the EV in the current and adjacent lanes are considered, with up to six SVs in total. Each SV state includes a presence flag, relative longitudinal and lateral positions, relative heading angle, and relative longitudinal and lateral speeds. The EV only considers SVs within the observation range $\Delta x \in [-80\,\mathrm{m},\,160\,\mathrm{m}]$.

\subsubsection{Reward Function}

The reward function is designed to consider driving safety, efficiency, and action consistency:
\begin{equation}
\label{reward_l}
\begin{split}
\begin{array}{l}
{{\cal R}^l}\! =\! {{\cal R}_{s}} + {{\cal R}_{e}} + {{\cal R}_{c}},
{{\cal R}_s}\! =\!  - 10{f_v} - 5\left( {{{\cal K}^h} + {{\cal K}^l}} \right),\\
{{\cal R}_e} = {{\left| {{v_e} - {v_*}} \right|} \mathord{\left/ {\vphantom {{\left| {{v_e} - {v_*}} \right|} {{v_*}}}} \right. \kern-\nulldelimiterspace} {{v_*}}} - \max \left( {0,{{{v_p} - {v_e}} \mathord{\left/ {\vphantom {{{v_p} - {v_e}} {{v_p}}}} \right. \kern-\nulldelimiterspace} {{v_p}}}} \right),\\
{{\cal R}_c}\! = \! -\! {{\left( {0.5\!\left| {{\delta _e}} \right|\! +\! 0.2\!\left| {\Delta {\delta _e}} \right|} \right)}}\! - \!{{\left( {0.5\!\left| {{a_e}} \right|\! +\! 0.2\!\left| {\Delta {a_e}} \right|} \right)}}
 \end{array}
\end{split}
\end{equation}
The weights for each term reflect the intended driving behavior: safety first, efficiency, and action consistency as a secondary objective. In safety reward ${{\cal R}_s}$, $f_v = 1$ indicates an EV violation, such as road departure or collision with SVs. The results from the safety evaluation module are also included. In efficiency reward ${{\cal R}_e}$, the EV is encouraged to reach the target speed $v_*$, while speeds below the threshold $v_p$ are penalized. We set $v_*$ and $v_p$ to $18m/s$ and $5m/s$, respectively. In action consistency reward ${{\cal R}_c}$, fluctuations in control commands are penalized to promote smoothness.

\subsection{Safety Mechanism Formulation}\label{sec:Implementation:B}

\subsubsection{Risk Severity Evaluation}

Artificial Potential Fields (APF) integrate discrete events into a unified field over high-dimensional observations \cite{2024TITS_chen}, providing reliable and expressive measures of driving risk severity. Based on APF, we propose a risk severity evaluation model for $\cal G$:
\begin{equation}
\label{K_APF}
\begin{split}
{\cal K} = \frac{1}{g}\sum\limits_{j = 1}^g {\left[ {{{\cal I}_j} \cdot \mathop {\max }\limits_{k \in \left[ {1, \cdots ,6} \right]} \left\{ {\rho _j^k} \right\}} \right]},
\end{split}
\end{equation}
where ${\rho _j^k}$ is the risk potential field at the $j$-th point relative to the $k$-th SV. Since points closer to the EV are more critical, the importance at the $i$-th point is: ${{\cal I}_j} = 1 - {e^{{K_r}\left( {j - g} \right)}}$, where $K_r$ is a decay rate coefficient. Additionally, ${\rho _j^k}$ is defined as:
\begin{equation}
\label{rho^k}
\begin{split}
\begin{array}{l}
\rho _j^k = {w_1}{e^{\left( { - \frac{1}{2}{{\rm P}_1}{{\rm B}^{ - 1}}{\rm P}_1^{\rm{T}}} \right)}} + {w_2}{e^{\left( { - \frac{1}{2}{{\rm P}_2}{{\rm B}^{ - 1}}{\rm P}_2^{\rm{T}}} \right)}}\\
{\rm B} = \left[ {\begin{array}{*{20}{c}}
{X_s^2}&0\\
0&{Y_s^2}
\end{array}} \right],{{\rm{P}}_1} = \left[ {\begin{array}{*{20}{c}}
{\Delta x_j^k}&{\Delta y_j^k}
\end{array}} \right],\\
{{\rm{P}}_2} = \left[ {\begin{array}{*{20}{c}}
{{_{\Delta a_x^k < 0}}\Delta x_j^k}&{{_{\Delta a_y^k < 0}}\Delta y_j^k}
\end{array}} \right]
\end{array}
\end{split}
\end{equation}
where ${w_1} \in [0.5,1]$ and ${w_1} + {w_2} = 1$. In addition, ${X_s}$ and ${Y_s}$ are the minimum safe distances in the longitudinal and lateral directions, respectively. The $\Delta x_j^k$ and $\Delta y_j^k$ are the longitudinal and lateral distances of the $j$-th point relative to the $k$-th SV.

\subsubsection{Prior Control Model Design}

To obtain $a^{l,r}$ and simplify the design, we combine the widely used Intelligent Driver Model (IDM) and Stanley path-tracking algorithm as a conservative prior control model. Specifically, IDM determines the acceleration based on environment, while Stanley algorithm computes the steering angle according to motion guidance.

\section{Experiments}\label{sec:Experiments}

\subsection{General Settings}

\subsubsection{Environment}

The training scenario is built in Highway-Env with three defined lanes \cite{2018highway_env}. At the start of each episode, the EV and SVs are randomly placed in any lane with random initial speeds. SVs follow the IDM and MOBIL models, allowing lane changes to reach their target speeds, which may interfere with the EV. All vehicles are modeled using the Kinematic Bicycle Model~\cite{car_model}. The vehicle capacity (V/C) ratio, representing traffic density. 

With this setup, training is conducted for 2,000 episodes using five seeds. Testing is then performed over 100 episodes on both Highway-Env and the HighD dataset \cite{2018HighD}, with each episode limited to 100s. Notably, the traffic density in Highway-Env is set to 0.3, presenting a challenging scenario.

\subsubsection{Comparison Methods}

We select several popular RL-based AD methods as baselines, which have different policy structures. \textbf{PPO}, serving as a low-level-only baseline, directly outputs control commands without any hierarchical guidance. The General Hierarchical method (\textbf{Gen-H-RL}, used as a high-level-only baseline) employs a high-level RL policy to select discrete behaviors for a rule-based low-level controller. Classical Hierarchical RL (\textbf{Class-HRL}) combines a high-level value-based policy with a low-level actor-critic structure. \textbf{RL-PTA} has an implicit hierarchical policy structure with a fixed timescale. \textbf{MHRL-I} and \textbf{Skill-Critic} are preliminary methods exploring multi-timescale RL policy training, with discrete and continuous high-level outputs, respectively. In contrast, our method features a more advanced hierarchical policy structure, in which P-AC is used to generate \textbf{H}ybrid action-based motion guidance (path points), with a supporting \textbf{S}afety mechanism included. The details of all methods are shown in Table~\ref{methods}. For all MT-based methods, the high- and low-level outputs operate at 1s and 0.1s, respectively, as this has proven effective~\cite{2023arXiv_wang}.

\begin{table}[!tb] 
\caption{Comparison of All Methods}
\vspace{-2mm}
\label{methods}
\centering
\begin{tabular}{ccccc}
\hline
\multirow{2}{*}{Method} & \multicolumn{2}{c}{High-Level} & \multirow{2}{*}{Low-Level} & \multirow{2}{*}{Is MT} \\
       & Model & Output & & \\
\hline
PPO \cite{2021ITSC_Agarwal} & N/A & N/A & PPO & No \\
Gen-H-RL \cite{2024ICDE_xia} & VB & Dis-Beh. & PID & Yes \\
Class-HRL \cite{2022TITS_Peng} & VB & Dis-Beh. & AC & No \\
RL-PTA \cite{OurITSC} & P-AC & Path Points & P-AC & No \\
MHRL-I \cite{2023Network_chen} & VB & Dis-Beh. & AC & Yes \\
Skill-Critic \cite{2024RAL_Hao} & AC & Path Points & AC & Yes \\
\begin{tabular}[c]{@{}c@{}}MTHRL-H (Ours)\end{tabular} & P-AC & Path Points& AC & Yes \\
\begin{tabular}[c]{@{}c@{}}MTHRL-HS (Ours)\end{tabular} & P-AC & Path Points & AC & Yes \\
\hline
\end{tabular}
\vspace{-4mm}
\end{table}

\subsubsection{Evaluation Metrics}

To comprehensively evaluate the performance of each driving policy, we select key metrics across four aspects:
\begin{itemize}
\item{Overall Performance: Total Reward (\textbf{TR}) per episode.}
\item{Efficiency: 1) Driving Speed (\textbf{DS}); 2) Total Lane Changes (\textbf{TLC}) per episode.}
\item{Action Consistency: 1) Absolute Steering angle (\textbf{AS}); 2) Absolute Acceleration (\textbf{AA}); 3) Centerline Departure Distance (\textbf{CDD}).}
\item{Safety: 1) Collision Rate (\textbf{CR}); 2) TTC in EV’s Current lane (\textbf{TTC-C}); 3) TTC in EV’s Target lane (\textbf{TTC-T}). TTC-C and TTC-T are identical if no lane change occurs.}
\end{itemize}

\subsection{Performance Comparison}

\subsubsection{Training}

\begin{figure}[!tb]
  \centering
  \includegraphics[width=0.47\textwidth]{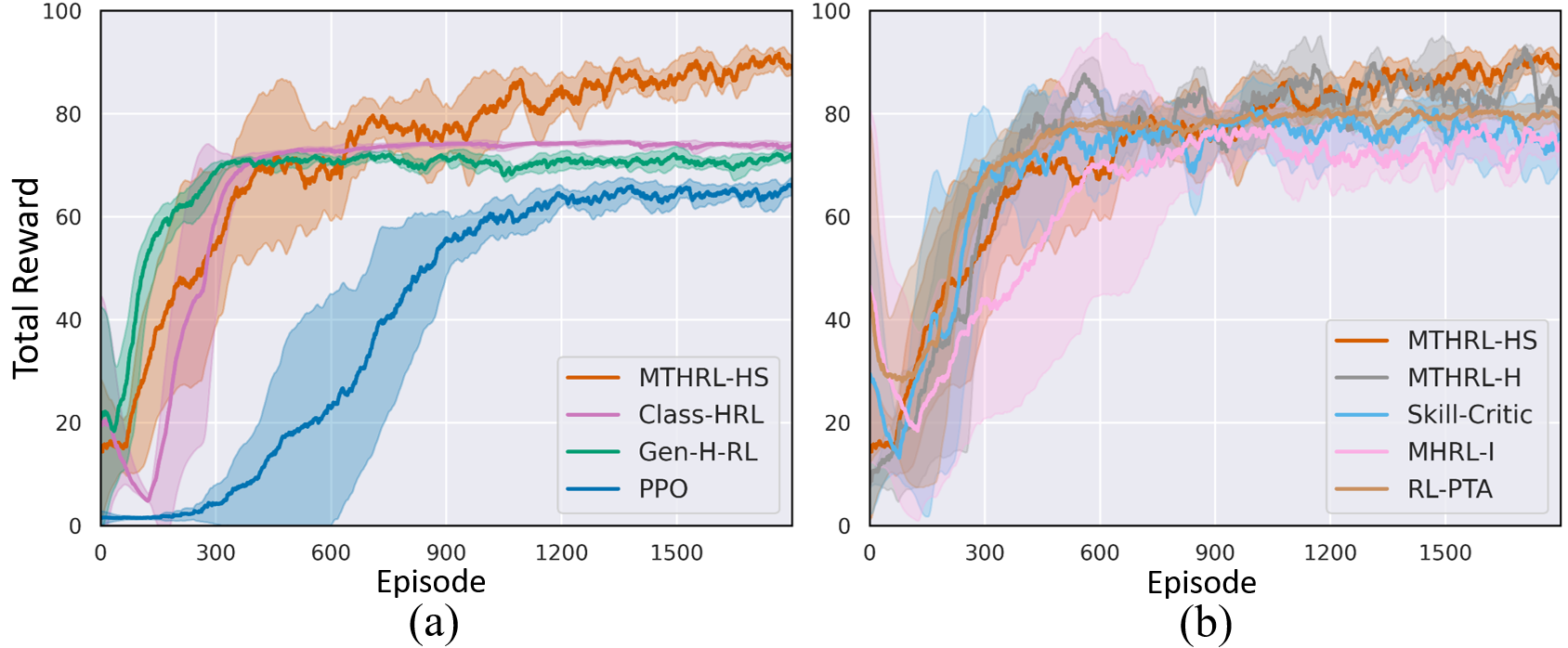}
  \vspace{-2mm}
  \caption{The training process of our method with comparison methods.}
  \label{fig:train}
    \vspace{-2mm}
\end{figure}

The total reward curves for all methods during training are shown in Fig.~\ref{fig:train}. All methods converge after 1,800 episodes. Our proposed MTHRL-HS achieves the highest reward, indicating superior driving performance.

In Fig.~\ref{fig:train}(a), PPO converges more slowly and achieve significantly lower rewards than methods using hierarchical techniques, suggesting that direct control output hinders effective policy learning. Gen-H-RL, which applies RL only at the high level, converges fastest but yields relatively lower rewards. In contrast, Class-HRL uses RL at both levels, leading to marginally improved performance yet slightly slower convergence. The methods in Fig.~\ref{fig:train}(b) generally benefit from more advanced hierarchical structures, resulting in better policies. RL-PTA, while implicitly hierarchical with a fixed timescale, achieves strong performance with less fluctuation across different seeds. MHRL-I/Skill-Critic perform significantly worse than MTHRL-H, demonstrating that hybrid action-based motion guidance, which better aligns with road structure, leads to superior policies compared to purely continuous or discrete actions. Furthermore, MTHRL-HS, with its safety mechanism, further enhances policy performance.

\subsubsection{Testing}

\begin{table*}[!tb]
\caption{Key indicators of test results in the Highway-Env simulation environment}
\vspace{-2mm}
\label{table:test data}
\begin{center}
\begin{tabular}{cccccccccc}
\toprule
\multirow{2}{*}{\textbf{Method}}                            & \textbf{Overall Perf.} & \multicolumn{2}{c}{\textbf{Efficiency}} & \multicolumn{3}{c}{\textbf{Action Consistency}}                       & \multicolumn{3}{c}{\textbf{Safety}}                       \\
\cmidrule(lr){2-2} \cmidrule(lr){3-4} \cmidrule(lr){5-7} \cmidrule(lr){8-10} 
                & \textbf{TR}                  & \textbf{DS} {[}$m/s${]}   & \textbf{TLC}  & \textbf{AS} {[}$rad${]} & \textbf{AA} {[}$m/s^2${]} & \textbf{CDD} {[}$m${]} & \textbf{CR} & \textbf{TTC-T} {[}$s${]} & \textbf{TTC-C} {[}$s${]} \\
\midrule
PPO        & 68.22(2.10) & 8.87(2.20)  & 1.11(0.93) & 0.086(0.184) & 0.98(0.68) & 0.22(0.35) & 0.01\% & 9.14(1.85) & 9.19(2.03) \\
Gen-H-RL   & 72.65(1.94) & 10.43(3.41) & 6.22(1.98) & \underline{0.022(0.090)} & \underline{0.42(0.63)} & \underline{0.09(0.24)} & 0.33\% & 7.67(1.93) & 7.13(2.38) \\
Class-HRL  & 73.20(0.58) & 9.99(3.16)  & 5.68(1.85) & 0.044(0.133) & 0.59(0.55) & 0.12(0.29) & 0.29\% & 7.42(2.18) & 7.51(2.10) \\
RL-PTA     & \underline{81.37(2.70)} & \underline{11.21(3.02)} & \underline{7.89(2.55)} & 0.024(0.052) & 0.54(0.51) & 0.13(0.25) & 0.10\% & \underline{9.10(1.19)} & 8.87(1.50) \\
MHRL-I   & 78.91(4.99) & 11.03(3.60) & 7.62(3.10) & 0.039(0.069) & 0.49(0.48) & 0.12(0.19) & \underline{0.09\%} & 9.05(1.26) & \underline{8.93(1.41)} \\
Skill-Critic   & 79.26(4.01) & 11.15(3.79) & 7.15(2.54) & 0.023(0.042) & 0.50(0.44) & 0.10(0.20) & 0.11\% & 8.89(1.87) & 8.73(1.95) \\
MTHRL-H(ours)   & 84.99(3.04)\textbf & \textbf{12.68(2.87)} & \textbf{8.19(2.69)} & \textbf{0.012(0.040)} & \textbf{0.31(0.37)} & \textbf{0.07(0.14)} & 0.07\% & 9.14(1.36) & \textbf{9.15(2.05)} \\
MTHRL-HS(ours) & \textbf{89.14(2.85)} & 12.61(3.81) & 8.03(2.60) & 0.013(0.034) & 0.33(0.36) & 0.07(0.16) & \textbf{0.03\%} & \textbf{9.36(1.26)} & 9.03(1.89) \\
\bottomrule
\end{tabular}
\end{center}
\vspace{-4mm}
\end{table*}

\begin{table*}[!tb]
\caption{Key indicators of test results in the HighD Dataset}
\vspace{-2mm}
\label{table:highd data}
\begin{center}
\begin{threeparttable}

\begin{tabular}{cccccccccc}
\toprule
\multirow{2}{*}{\textbf{Method}}                            & \textbf{Overall Perf.} & \multicolumn{2}{c}{\textbf{Efficiency}} & \multicolumn{3}{c}{\textbf{Action Consistency}}                       & \multicolumn{3}{c}{\textbf{Safety}}                       \\
\cmidrule(lr){2-2} \cmidrule(lr){3-4} \cmidrule(lr){5-7} \cmidrule(lr){8-10} 
                & \textbf{TR}                  & \textbf{DS} {[}$m/s${]}   & \textbf{TLC}  & \textbf{AS} {[}$rad${]} & \textbf{AA} {[}$m/s^2${]} & \textbf{CDD} {[}$m${]} & \textbf{CR} & \textbf{TTC-T} {[}$s${]} & \textbf{TTC-C} {[}$s${]} \\
\midrule
PPO        & 81.62(3.54) & 10.22(1.92) & 0.68(1.20) & 0.081(0.188) & 1.08(0.65) & 0.18(0.23) & 0.02\% & 9.23(1.88) & 9.12(1.68)\\
Gen-H-RL   & 87.40(5.33) & 13.08(2.40) & 5.73(2.14) & 0.030(0.073) & \underline{0.36(0.51)} & \underline{0.07(0.18)} & 0.20\% & 8.06(1.85) & 7.66(2.00) \\
Class-HRL  & 89.25(3.11) & 12.82(3.15) & 5.55(2.16) & 0.041(0.121) & 0.74(0.58) & 0.11(0.20) & 0.21\% & 7.84(1.99) & 7.89(2.07) \\
RL-PTA     & \underline{95.57(7.04)} & \underline{15.01(2.98)} & 6.96(3.84) & 0.022(0.050) & 0.60(0.52) & 0.12(0.23) & 0.05\% & \underline{9.21(1.15)} & 8.99(1.43) \\
MHRL-I   & 94.91(8.95) & 14.13(3.63) & \underline{7.42(4.23)} & 0.028(0.062) & 0.53(0.50) & 0.12(0.18) & 0.08\% & 9.10(1.21) & \underline{9.04(1.28)} \\
Skill-Critic   & 95.30(8.01) & 14.80(4.04) & 7.08(3.80) & \underline{0.021(0.039)} & 0.59(0.39) & 0.10(0.21) & \underline{0.05\%} & 8.96(1.77) & 8.97(1.81) \\
MTHRL-H(ours)   & 96.62(6.30) & \textbf{16.18(3.00)} & \textbf{7.39(3.52)} & 0.013(0.041) & 0.37(0.35) & \textbf{0.06(0.15)} & 0.05\% & 9.15(1.76) & \textbf{9.10(1.88)} \\
MTHRL-HS(ours) & \textbf{96.77(5.51)} & 16.09(3.44) & 7.07(3.27) & \textbf{0.013(0.037)} & \textbf{0.32(0.30)} & 0.07(0.17) & \textbf{0.02\%} & \textbf{9.30(1.33)} & 9.09(1.72) \\
\bottomrule
\end{tabular}
\begin{tablenotes}
\footnotesize
\item[*] Boldface indicates the best result among all methods; underline indicates the best result among baseline methods.
\end{tablenotes}
\end{threeparttable}
\end{center}
\vspace{-4mm}
\end{table*}

\begin{figure}[!tb]
  \centering
  \includegraphics[width=0.47\textwidth]{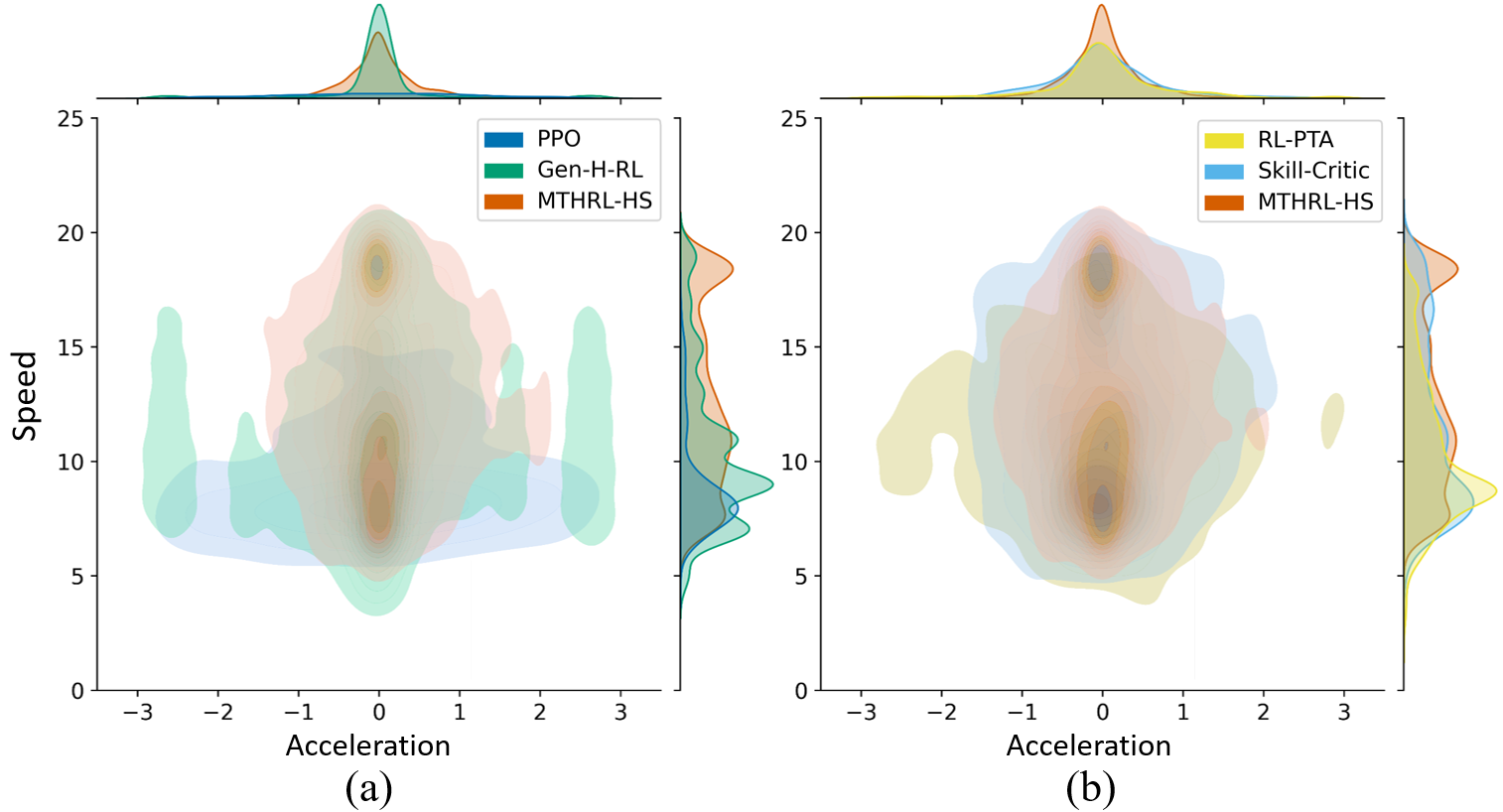}
  \vspace{-2mm}
  \caption{The joint distribution of acceleration and speed in testing.}
  \label{fig:joint_acc}
    \vspace{-2mm}
\end{figure}

\begin{figure}[!tb]
  \centering
  \includegraphics[width=0.47\textwidth]{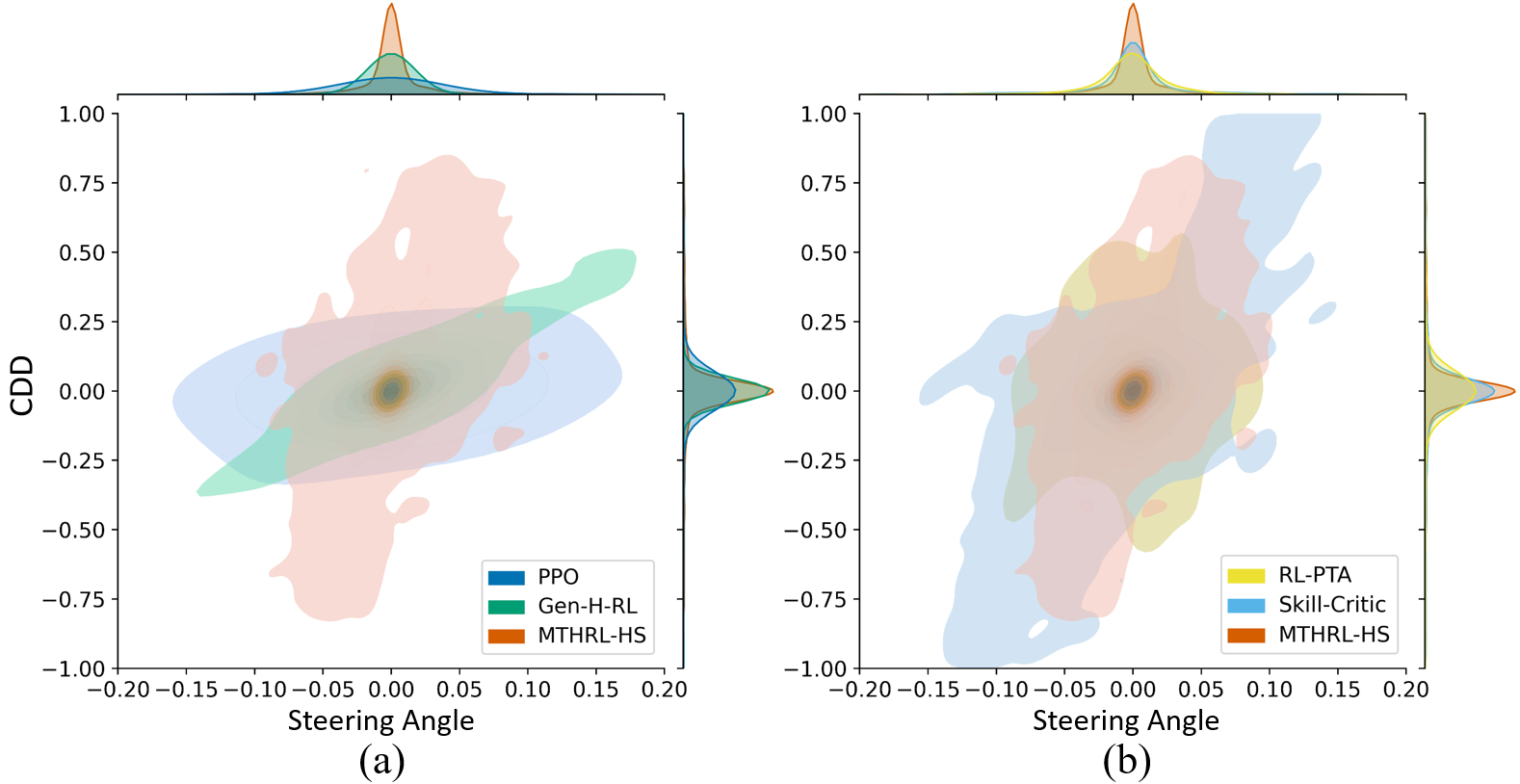}
  \vspace{-2mm}
  \caption{The joint distribution of steering angle and CDD in testing.}
  \label{fig:joint_steer}
    \vspace{-2mm}
\end{figure}

\begin{figure}[!tb]
  \centering
  \includegraphics[width=0.47\textwidth]{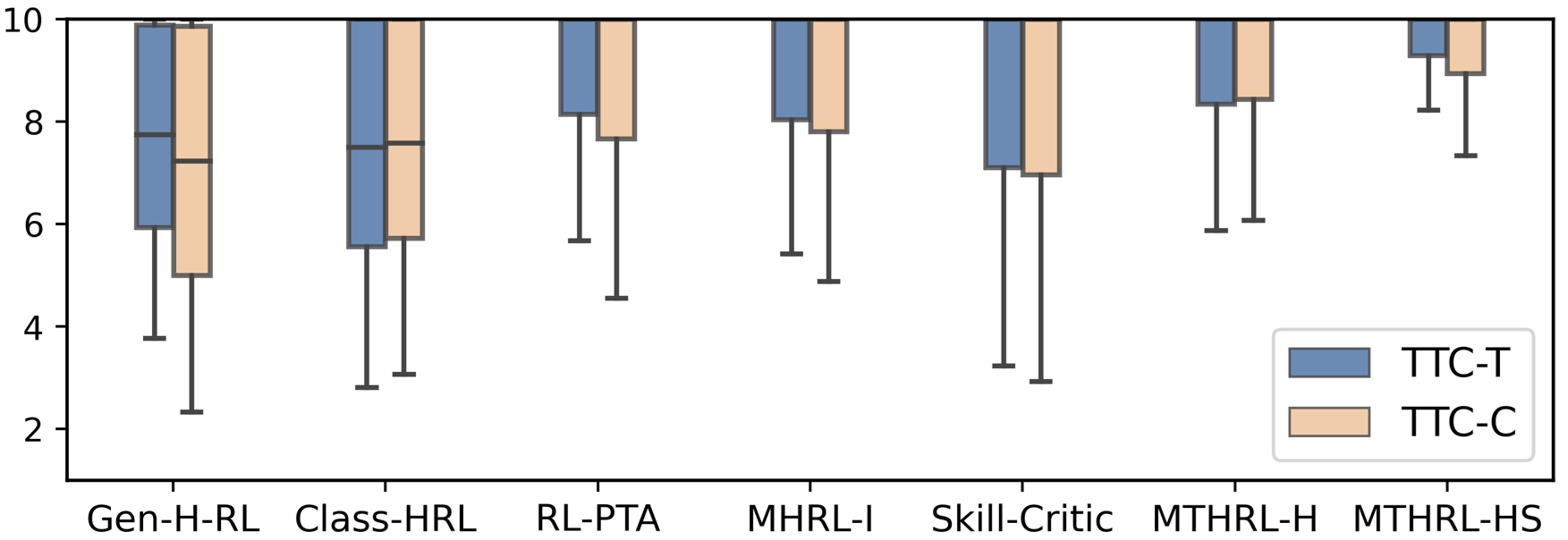}
  \vspace{-2mm}
  \caption{The distributions of TTC-C and TTC-T in testing.}
  \label{fig:TTC_box}
    \vspace{-2mm}
\end{figure}


The testing results further show that different methods produce distinct driving behaviors, as detailed in Table~\ref{table:test data}. Notably, our method achieves the highest TR, consistent with the training results. Compared to RL-PTA, which has the second-highest TR, MTHRL-H improves TR by 4.4\%, and with the safety mechanism ‘S’, this improvement increases to 9.5\%. After analyzing all the metrics, we provide additional details for some key metrics of the well-performing methods. The joint distribution of acceleration and speed is shown in Fig.~\ref{fig:joint_acc}, while the joint distribution of steering angle and CDD is shown in Fig.~\ref{fig:joint_steer}. The Fig.~\ref{fig:TTC_box} further presents the distributions of TTC-C and TTC-T as boxplots.

For driving efficiency, MTHRL-H achieves the highest DS and TLC, with DS improved by 13.1\% over the suboptimal method. This indicates that the multi-timescale hierarchical policy enables more flexible lane-changing behaviors to enhance driving efficiency, while the introduction of safety mechanism does not significantly compromise this performance. More specifically, under the same testing environment, Fig.~\ref{fig:joint_acc} shows that the peak speed distributions for each method correspond to two potential cases: one near 20 m/s, representing opportunities for overtaking to reach the target speed; and another near 10 m/s, indicating the EV must follow SVs due to traffic congestion. Notably, MTHRL-HS achieves better lane-change timing in the former case, yielding the highest peak speed distribution. In the latter case, it effectively chooses to follow a faster SV, shifting the peak speed distribution upward. In contrast, other methods show lower driving efficiency, with worse DS, TLC, and speed distributions. In particular, PPO, lacking hierarchical structures, tend to adopt overly conservative following policies.

For driving action consistency, MTHRL-H shows the lowest means and standard deviations for AS, AA, and CDD, which are 50.0\%, 26.2\%, and 22.2\% lower than those of the suboptimal method, respectively. The safety mechanism does not adversely affect these metrics. As shown in Fig.~\ref{fig:joint_acc} and Fig.~\ref{fig:joint_steer}, our method produces acceleration, steering angle, and CDD values more tightly centered around 0. Even during lane changes, while CDD increases, steering angles remain smaller than with other methods. This suggests that high-level hybrid action-based motion guidance improves action consistency. It reduces fluctuations in control commands, resulting in smoother longitudinal and lateral driving behaviors. In contrast, PPO, which directly outputs control commands, lead to more fluctuating behaviors and make it harder to keep the EV on the centerline. Among hierarchical methods, generating finer-grained path points at the high level yields greater action consistency than discrete behavior generation.

For driving safety, excluding the over-conservative PPO, MTHRL-H reduces the CR by 22.2\% compared to the suboptimal method. With the safety mechanism, the reduction in CR increases to 66.7\%, clearly demonstrating its effectiveness in improving safety. Additionally, TTC-C and TTC-T reflect the policy’s ability in ensuring safer driving during interactions with SVs. As shown in Table~\ref{table:test data} and Fig.~\ref{fig:TTC_box}, MTHRL-HS achieves the highest TTC-C and TTC-T, indicating a more cautious driving style. Meanwhile, the safety mechanism results in a significantly higher TTC-T than TTC-C, highlighting its ability to guide the EV into safer lanes.

\subsubsection{Validation on HighD Dataset}

The metrics in Table~\ref{table:highd data} show the driving performance of each method on the HighD dataset. Compared to the Highway-Env scenario, all methods perform better due to lower traffic density in HighD. MTHRL-H and MTHRL-HS remain the top performers overall, despite a smaller TR gap with other methods, demonstrating strong adaptability and robust policy performance in real traffic. Benefiting from the multi-timescale hierarchical architecture and hybrid-action-based motion guidance, MTHRL-HS and MTHRL-H maintain more efficient and stable driving policies. Unlike testing in Highway-Env, MTHRL-HS performs nearly similar to MTHRL-H in action consistency, even slightly outperforming it. For safety, excluding the over-conservative PPO, MTHRL-HS further reduces CR to 0.02\% through hierarchical safety mechanism. This indicates that the safety mechanism not only enhances safety but also enables RL policy to maintain more consistent actions when encountering unfamiliar SVs. Therefore, above results confirm the superiority of our method across all driving metrics and its strong potential for real-world deployment. Notably, since real deployment is safety-critical, MTHRL-HS is the preferred choice, as it significantly enhances safety with only minor efficiency loss.

\section{Conclusion and Future Work}\label{sec:Conclusion}

This paper proposes a Multi-Timescale Hierarchical RL approach for AD. The approach features a hierarchical policy structure: a high-level RL policy generates long-timescale motion guidance, while a low-level RL policy produces short-timescale vehicle control commands. Therein, a hybrid action-based explicit representation is designed for motion guidance to better adapt to structured roads and to facilitate addressing low-level state inconsistencies. In addition, supporting hierarchical safety mechanisms are introduced to enhance the safety of both high- and low-level outputs. We evaluate our approach against advanced baselines in both simulator-based and HighD data-based highway multi-lane scenarios, and conduct a comprehensive analysis of various driving behavior metrics. Results demonstrate that our approach effectively improves driving efficiency, action consistency, and safety.

Future work aims to: i) extend our approach to more~complex scenarios such as bidirectional roads and intersections, introducing generalization techniques as necessary; ii) design advanced safety mechanisms with comparisons to safety‑oriented baselines; and iii) incorporate higher‑fidelity vehicle dynamics models toward deployment in real vehicular systems.


\bibliographystyle{ieeetr}
\bibliography{reference}

\end{document}